\documentclass[12pt]{article}
\usepackage{frExamplee}
\usepackage{graphicx}
\usepackage{natbib}
\usepackage{setspace}
\usepackage[all]{nowidow}
\usepackage[caption=false,font=footnotesize,labelfont=sf,textfont=sf]{subfig}
\usepackage[export]{adjustbox}
\usepackage{cite}
\usepackage{multicol}
\usepackage[bookmarks=true]{hyperref}
\usepackage{array}
\usepackage{times}
\usepackage{xcolor}
\usepackage{multirow}
\usepackage{hhline}

\newcommand{\fig}[1]{Fig.~\ref{#1}}
\newcommand{\sect}[1]{Sec.~\ref{#1}}
\newcommand{\tab}[1]{Table~\ref{#1}}

\newcommand{\unit}[2]{{#1}\,{#2}}

\begin{document}

\title{Providentia -- A Large-Scale Sensor System for the Assistance of Autonomous Vehicles and Its Evaluation}

\author{
Annkathrin~Kr{\"a}mmer\thanks{These authors contributed equally to this work.} \\
fortiss GmbH\\
Munich, Germany\\
\And
Christoph~Sch{\"o}ller$^{*}$\\
fortiss GmbH\\
Munich, Germany\\
\AND
Dhiraj~Gulati\\
Mentor Graphics\\
Munich, Germany\\
\And
Venkatnarayanan~Lakshminarasimhan\\
Technical University of Munich\\
Munich, Germany\\
\And
Franz~Kurz\\
German Aerospace Center (DLR)\\
Wessling, Germany\\
\And
Dominik~Rosenbaum\\
German Aerospace Center (DLR)\\
Wessling, Germany\\
\And
Claus~Lenz\\
Cognition Factory GmbH\\
Munich, Germany\\
\And
Alois~Knoll\\
Technical University of Munich\\
Munich, Germany\\
}

\maketitle
\vspace{-0.5cm}
\begin{abstract}
\vspace{-0.5 cm}The environmental perception of an autonomous vehicle is limited by its physical sensor ranges and algorithmic performance, as well as by occlusions that degrade its understanding of an ongoing traffic situation. This not only poses a significant threat to safety and limits driving speeds, but it can also lead to inconvenient maneuvers. Intelligent Infrastructure Systems can help to alleviate these problems. An Intelligent Infrastructure System can fill in the gaps in a vehicle's perception and extend its field of view by providing additional detailed information about its surroundings, in the form of a digital model of the current traffic situation, i.e. a digital twin. However, detailed descriptions of such systems and working prototypes demonstrating their feasibility are scarce. In this paper, we propose a hardware and software architecture that enables such a reliable Intelligent Infrastructure System to be built. We have implemented this system in the real world and demonstrate its ability to create an accurate digital twin of an extended highway stretch, thus enhancing an autonomous vehicle's perception beyond the limits of its on-board sensors. Furthermore, we evaluate the accuracy and reliability of the digital twin by using aerial images and earth observation methods for generating ground truth data.
\end{abstract}

\section{Introduction}
\label{sec:introduction}
The environmental perception and resulting scene and situation understanding of an autonomous vehicle are limited by the available sensor ranges and object detection performance. Even in the vicinity of the vehicle, the existence of occlusions leads to incomplete information about its environment. The resulting uncertainties pose a safety threat not only to the autonomous vehicle itself but also to other road users. To enable it to operate safely, it is necessary to reduce its driving speed, which in turn slows down traffic. Furthermore, this incomplete information results in impaired driving comfort, as the vehicle must spontaneously react to unforeseen scenarios.

\begin{figure}[t!]
\centering
\includegraphics[trim=0 0 0 10,clip, width=0.9\columnwidth]{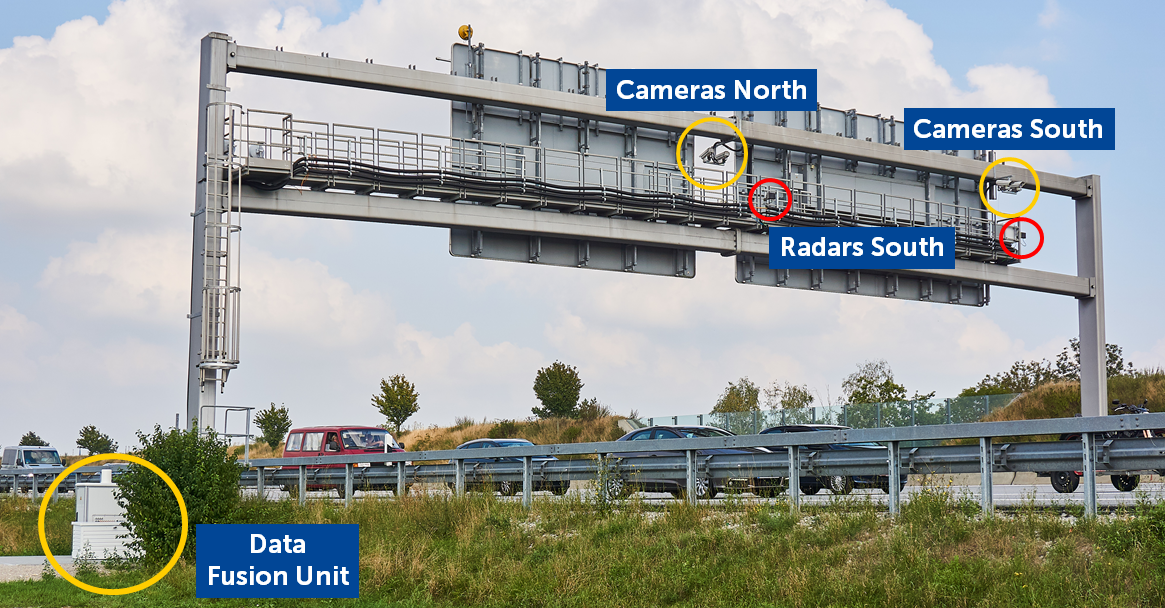}
\caption{One of the Providentia measurement points on the A9 highway. The two radars directed towards the north are installed on the other side of the gantry bridge and are therefore not visible from this perspective.}
\label{fig:measurement-point}
\end{figure}

Intelligent Infrastructure Systems (IIS) can alleviate these problems by providing autonomous vehicles -- as well as conventional vehicles and drivers -- at operating time with complementing information about each road user and the overall traffic situation~\citep{qureshi2013survey, menouar2017uav}, thereby greatly extending their perception range as well. In particular, an IIS can observe and detect all road users from multiple superior perspectives, with extended coverage compared to that of an individual vehicle. Providing a vehicle with this additional information gives it a better and spatially extended understanding of its surrounding scene and enables it to plan its maneuvers more safely and proactively. Furthermore, an IIS with the described capabilities enables a multitude of services that further support decision making.

However, actually building such a system involves a number of challenges, such as the right choice of hardware and sensors, and their optimal deployment and utilization in a complex software stack. Its perception must remain reliable and robust in a wide variety of weather, light and traffic conditions. Ensuring such reliability necessitates a combination of multimodal sensors, redundant road coverage with overlapping field of views (FoV), accurate calibration~\citep{schoeller2019radarcalib}, and robust detection and data fusion algorithms.

Since we sketched ideas about how such a system could be designed in previous work~\citep{hinz2017designing}, in this paper we propose a concrete, scalable architecture. This architecture is the result of the experience we made with the real world build-up of the IIS Providentia~(see \fig{fig:measurement-point}). It includes the system's hardware as well as the software to operate it. In terms of hardware, we discuss the choice of sensors, the network architecture and the deployment of edge computing devices to enable fast and distributed processing of heavy sensor loads. We outline our software stack and the detection and fusion algorithms used to generate an accurate and consistent model of the world, which we call the digital twin. The digital twin includes information such as position, velocity, vehicle type and a unique identifier for every observed vehicle. By providing this digital twin to an autonomous driving research vehicle, we demonstrate that it can be used to extend the limits of the vehicle's perception far beyond its on-board sensors.

For autonomous vehicles to trust the digital twin for maneuver planning, its accuracy and reliability must be known. However, a thorough evaluation requires precise ground truth about the traffic situation. This is non-trivial to obtain. To solve this issue, we took aerial images of the traffic in our testbed to generate an approximate ground truth and use it to evaluate our system. While we explained the underlying idea in \citet{kraemmer2020vorausschauend}, in this paper we describe in detail the methods used for this evaluation. We present the results of our evaluation of the Providentia system and analyze the system's performance in real world applications. Our evaluation methodology is not specific to our system and can serve as a general framework for the evaluation of IIS.

\section{Related Work}
\label{sec:related-work}
First ideas for assisting vehicles, as well as monitoring and controlling traffic with an IIS have already been developed in the PATH~\citep{shladover1992california} and PROMETHEUS~\citep{braess1995prometheus} projects. Recently, with the growing efforts of industry and research to realize autonomous driving, the need for IIS that are able to support autonomous vehicles has further increased. Several new projects have therefore been initiated, with the goal of developing and researching on prototypical IIS. However, their focuses differ widely and few detailed system descriptions are available.

\textbf{Communication.} Some IIS projects primarily focus on the communication aspects between the vehicle and infrastructure, and sometimes additionally vehicle-to-vehicle communication. The research project \citet{diginetps} focuses in particular on the communication of traffic signal information, parking space occupancy, traffic density and road conditions to vehicles. Similarly, the \citet{veronika} project provides traffic signal information with the goal of reducing emissions and energy consumption. The \citet{antwerpsmart} test site is built along a \unit{4}{km} highway strip and equipped with road side communication units on gantry bridges. In contrast to our work, its research focus is on vehicle-to-everything communication and distributed edge computing. Similarly, the goal of the \citet{nycconnected} is to improve safety and reduce the number of crashes by providing drivers with alerts via Dedicated Short-Range Communication. The \citet{m-city} project built an artificial test facility to evaluate the performance of connected and automated vehicles. Research that uses this test facility is primarily on communication, but also partially covers roadside perception.

\textbf{Roadside Perception.} The primary goal of roadside perception systems is the enhancement of autonomous vehicle safety. The system in the Test Area Autonomous Driving Baden-W{\"u}rttemberg \citep{tafbw2018fleck} is perceiving a cross-road with two cameras and creates a digital twin. It also provides functionality to evaluate autonomous driving functions in a realistic environment. However, this system is much smaller than Providentia, and cannot operate at night as it only uses cameras. In the MEC-View project, an IIS consisting of cameras and lidars mounted on streetlights creates a real-time environment model of an urban intersection that is amongst others fused into a vehicle's on-board perception system~\citep{gabb2019infrastructure}. Furthermore, the local highway operator in Austria is transforming its existing road operator system into an IIS~\citep{seebacher2019infrastructure}. It is aiming at actively supporting autonomous vehicles and at enabling the validation of autonomous vehicle perception.
 
\textbf{IIS Algorithms.} Instead of the IIS itself, many research contributions propose methods of making algorithmic use of the information provided by an IIS, or optimizing their function. With regard to communication networks, \citet{miller2008v2v2i} proposes an architecture for efficient vehicle-to-vehicle and vehicle-to-infrastructure communication, while \citet{kabashkin2015bidirv2x} analyses the reliability of bidirectional vehicle-to-infrastructure communication. In the project \citet{kora9}, \citet{geissler2019optimize} formulate an optimization problem that maximizes sensor coverage to locate suitable sensor placements in an IIS. Popular areas of research in the field of computer vision that are related to IIS include traffic density prediction~\citep{zhang2017vehiclecount, zhang2017understanding} and vehicle re-identification~\citep{shen2017learning, zhou2018viewpoint}. Other topics involving information provided by an IIS include danger recognition~\citep{yu2018traffic} and vehicle motion prediction~\citep{diehl2019graph, liu2019integrated}.

In this paper, we focus on the overall system architecture and implementation of a large-scale IIS that generates a digital twin of the current traffic. The aim of our system is to complete and extend a vehicle's perception and to provide information that enables the implementation of various algorithms and applications based on the digital twin. In the literature, detailed technical descriptions of systems with similar size and capabilities like ours are not publicly available to the best of our knowledge.

Furthermore, to ensure our system's performance is suited for the intended purposes, we conduct a thorough evaluation by considering the overall traffic, rather than trajectories from a single test vehicle. Thereby we account for a broad variety of vehicle types, colors and driving behaviors. In particular, we evaluate the spatial accuracy as well as the detection rate, i.e. the system's performance with respect to missing vehicles and false detections. To the best of our knowledge, our work is the first to provide such quantitative results on the performance of a large-scale IIS for fine-grained vehicle perception.

\section{Providentia System Architecture}
\label{sec:architecture}
\begin{figure}
\centering
\includegraphics[trim=0 80 0 80,clip, width=1.00\columnwidth]{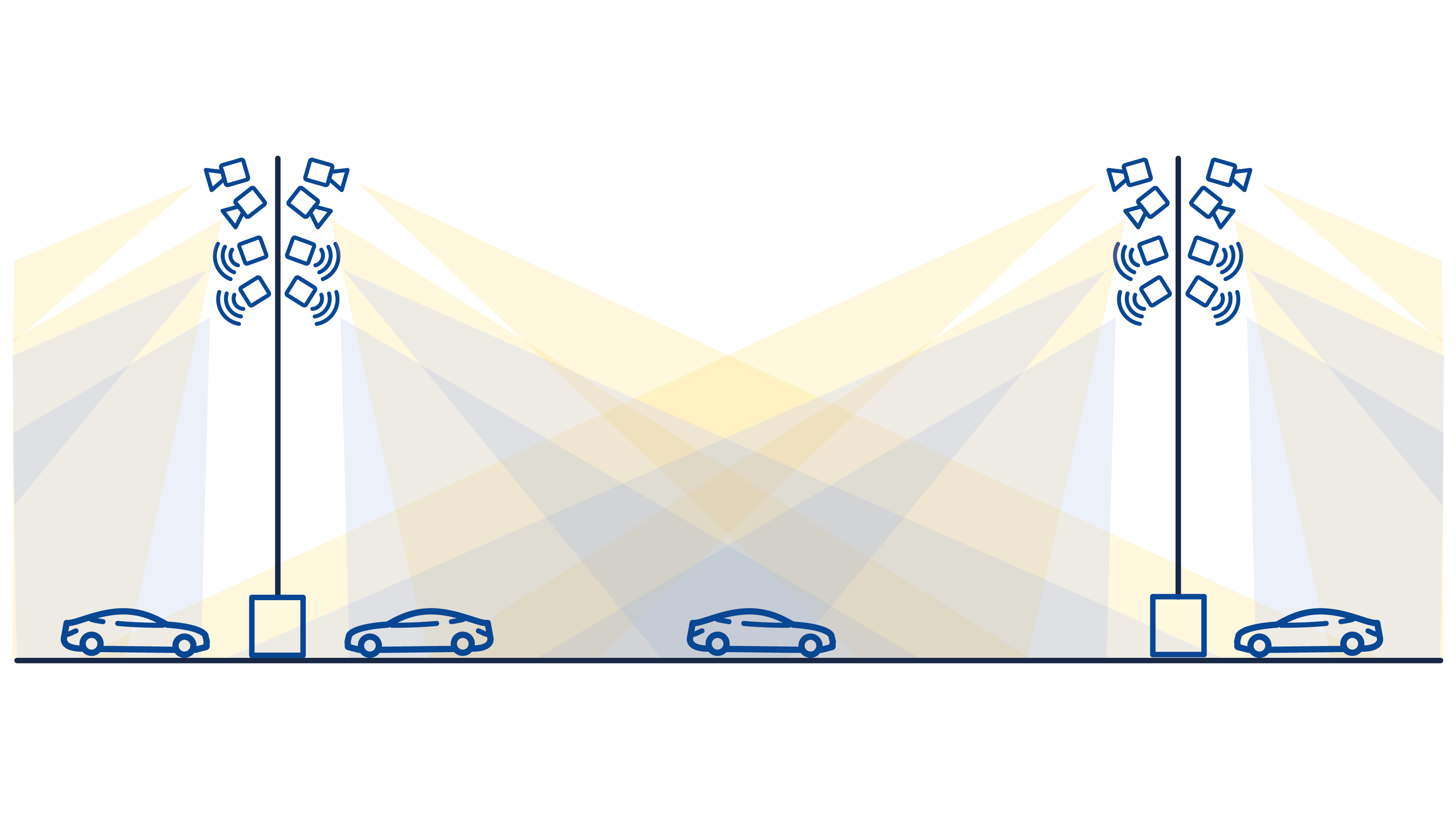}
\caption{Schematic illustration of the Providentia sensor setup, with overlapping FoVs for redundancy.}
\label{fig:redundant-sensor-setup}	
\vspace*{-3mm}
\end{figure}

In this section, we describe the design of the Providentia system, including the hardware and software setup and the algorithms used for detection, calibration, fusion and position refinement.

Providentia is a large-scale distributed sensor system consisting of multimodal sensors, multiple edge computing units, a complex software architecture, and a broad range of state-of-the-art algorithms. It is built along the A9 highway close to Munich. Its primary purpose is to provide a real-time and reliable digital twin of the current road traffic at any given time or day of the year, for use in a variety of applications. 

\subsection{Hardware and Software Setup}
At the time of writing, two gantry bridges~--~separated by a distance of approximately \unit{440}{m}~--~have been equipped with sensors and computing hardware. Each of these gantry bridges represents one measurement point in our system as illustrated in~\fig{fig:measurement-point}. To achieve high perception robustness, we use sensors of different measurement modalities that cover the entire stretch between our measurement points redundantly. \fig{fig:redundant-sensor-setup} illustrates the overall setting of the system with the redundant coverage of the highway.

Each measurement point comprises eight sensors with two cameras and two radars per viewing direction. In each direction, one radar covers the right-hand side while the other covers the left-hand side of the highway. The cameras have focal lengths of \unit{16}{mm} and \unit{50}{mm} to enable them to capture both the far and near ranges, while covering the entire width of the highway. By combining sensors with different measuring principles, our system is able to operate in varying traffic, light and weather conditions. Besides having redundant coverage with the sensors on each measurement point, we also selected the positions of the two measurement points in such a way that their overall FoVs overlap. This further increases redundancy and thus robustness, and allows smooth transitions while tracking vehicles as they move from one measurement point to the other. In addition, covering the highway stretch from different viewing directions helps to resolve detection errors and occlusions.

The system employs specialized \unit{24}{GHz} traffic monitoring radars from SmartMicro, of the generation \mbox{UMRR-0C}, with a type 40 antenna. They provide detections at an average frequency of \unit{13.2}{Hz}. They are specifically designed for stationary traffic monitoring and have a good object separation capability, even of closely spaced objects. Furthermore, they have a high detection range of up to \unit{350}{m} -- \unit{450}{m}, depending on the object size and driving direction. Each radar covers up to 256 objects on up to 7 lanes for the side of the highway it specializes on. All of these properties are necessary for traffic detection on high throughput highways.

The cameras are Basler \mbox{acA1920-50gc}, taking color images at an average frequency of \unit{25}{Hz}. After testing various other cameras, we selected this model especially because it can provide raw images with a very short processing time and hence very short latency, which is necessary for creating a real-time digital twin. The raw images allow us to define the image compression level ourselves, such that artifacts are minimized and our detection algorithms become as accurate as possible.

All the sensors at a single measurement point are connected to a Data Fusion Unit (DFU), which serves as a local edge computing unit and runs with Ubuntu 16.04 Server. It is equipped with two INTEL Xeon E5-2630v4 \unit{2.2}{GHz} CPUs with \unit{64}{GB} RAM and two NVIDIA Tesla V100 SXM2 GPUs. All sensor measurements from the cameras and radars are fed into the detection and data fusion toolchain running on this edge computing unit. This results in object lists containing all the road users tracked in the FoV of that measurement point. Each DFU transmits this object list to a backend machine via a fibre optic network, where they are finally fused into the digital twin that covers the entire observed highway stretch.

The full architecture is shown in \fig{fig:architecture}. We use ROS~\citep{quigley2009ros} on all computing units to ensure seamless connectivity. The final digital twin is communicated either to autonomous vehicles or to a frontend, where it can be visualized as required for drivers or operators.

\begin{figure*}[t]
\centering
\includegraphics[trim=0 0 0 0,clip, width=0.99\textwidth]{architecture}
\caption{Platform architecture of the Providentia system.}
\label{fig:architecture}	
\end{figure*}

\subsection{Object Detection}
\label{sec:object-detection}

The first step towards creating the digital twin of the highway is to detect and classify the vehicles on the highway. Our traffic radars are capable ex-factory of detecting objects and providing time-stamped positions and velocities in their respective local sensor coordinate systems on street level. We transform this output of each radar into our system's global Cartesian coordinate system using the radar calibration parameters in preparation for data fusion.

Detection and classification of objects in the camera images are performed by the DFU edge devices next to the highway. The system's cameras publish time-stamped images that are tagged with a unique camera identifier. To ensure scalability and safety even in the event of camera failures, our modular object detection pipelines subscribe to each image stream separately. The object detection pipelines are constantly monitored to supervise and analyze detection performance. The modular services are automatically restarted in the event of any failures. Not only are the multiple camera streams processed in parallel, the object detection can also work with various detection networks. This allows us to configure the object detection to optimally balance between low computation time and high accuracy, depending on the requirements that the application of our system poses. To this end, we performed extensive research on state-of-the-art detection algorithms, based on neural networks~\citep{altenberger2018non}.

At the time of writing, we have been using the YOLOv4~\citep{bochkovskiy2020yolov4} architecture as our detection network in the object detection pipelines. In addition to regressing two-dimensional bounding boxes with a confidence score, this network classifies the detected vehicles into types as car, truck, bus or motorcycle. The output is then published prior to transformation.

To compute the three-dimensional positions of the vehicles from the camera detections in the images, it is unfavorable to use stereo vision techniques in our system. Our cameras that look in the same driving direction are placed close together, significantly differ in their focal length, and their FoVs overlap only partially. Instead, we use the vehicles' bounding boxes to cast a ray through their lower-edge midpoint and intersect it with the street-level ground plane that we know from our camera calibration. We transform the resulting vehicle positions into our system's global Cartesian coordinate system in the same manner as the detections of the radars. All of the resulting measurements are then ready to be fused into a consistent world model and are fed into the data fusion pipeline, starting with a tracking module.

\subsection{Calibration}
\label{sec:calibration}

Precise calibration of each sensor and measurement point is necessary to enable the transformation of all sensor measurements from their respective local system to a common global coordinate system. Only then we can perform data fusion and ultimately generate the digital twin. As the first step, we intrinsically calibrated all cameras individually prior to their installation using a common checkerboard calibration target and the camera calibration package provided by ROS. In particular, the function we used is minimizing reprojection errors with the Levenberg-Marquardt optimization algorithm. During build-up of the system, all radars were calibrated with their \mbox{ex-factory} supplied calibration software. This software assists in ensuring that each radar is mounted with ideal orientation at its perfect operating point, such that the radar optimally covers the part of the highway it specializes on. To determine these parameters, the software requires the radar's mounting height and a local map as input. As a result, the radar is enabled to internally transform all measurements and to output them in a Cartesian coordinate system with its origin placed directly underneath the radar on street level. The $XY$-plane of this coordinate system approximates the street.

The overall extrinsic calibration of the system after having installed all sensors is non-trivial. Not only does our system possess a high number of sensors and degrees of freedom, but it also makes use of sensors with heterogeneous measurement principles. Once we calibrated and know all the sensors' positions and orientations on each gantry bridge, we can transform their measurements in a common global coordinate system. Our system can then provide the digital twin in a standardized reference frame to the outside world, as we know the measurement points' orientation towards north and the GPS coordinates of reference points on the gantry bridges from official surveying and a HD map.

To obtain approximate starting points for each sensor for our extrinsic calibration algorithms, we manually measure the relative translation of all sensors and the reference point on each gantry bridge separately using modern laser distance meters. Thereby, we measure orthogonal respectively parallel to the gantry bridge as well as the street plane for reference. Note that the gantry bridges are accessible with a walking corridor on top (see \fig{fig:measurement-point}), which allows taking manual measurements. While the radars' orientations are already approximately known from set-up, we determine all cameras' yaw angles relative to the driving direction with a compass, and their pitch and roll angles relative to the horizontal street plane with a digital angle finder with spirit levels.

\begin{figure}[t]
\centering
\subfloat[Initial Calibration]{\includegraphics[trim=0 20 0 0,clip,width=0.49\columnwidth]{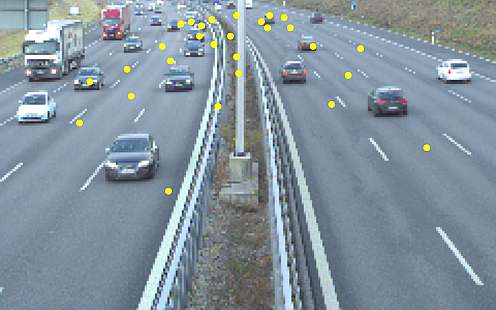}}
\hfill
\subfloat[Improved Calibration]{\includegraphics[trim=0 20 0 0,clip,width=0.49\columnwidth]{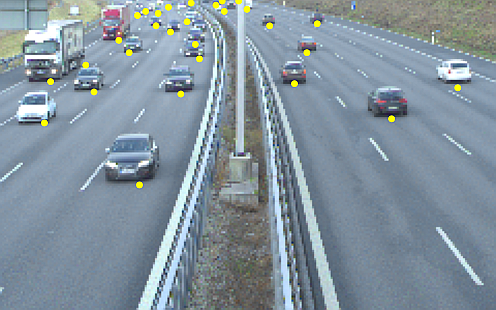}}
\caption{After our initial calibration with physical measurements (a), we refine the inter-sensor calibration by projecting the radar detections in the camera image and optimize the alignment with the observed vehicles (b).}
\label{fig:calib}
\vspace*{-3.5mm}
\end{figure}

We then refine the camera poses using vanishing point methods that utilize parallel road markings to find vanishing points~\citep{kanhere2010vanish}. To fine-tune the inter-sensor calibration, we manually minimize the projection and re-projection errors for all sensor pairs, both in the image planes (see \fig{fig:calib}) and on street level in the three-dimensional coordinate system. Furthermore, we incorporate the lane information from the HD map as an additional reference.
The final calibration step is to refine the alignment of the measurement points with each other. By transforming the detections of both measurement points into the same coordinate system, we are able to manually associate them and minimize their distance to find an optimal overall calibration. This results in a global coordinate system for our IIS, where all sensor detections can be transformed into. In this coordinate system, the digital twin can be created and then transformed to GPS coordinates.

\subsection{Data Fusion}
\label{sec:fusion}

When it comes to the sensor data fusion, a large-scale system such as ours poses many challenges. On the highway, we can observe a very large number of vehicles that have to be tracked in real time. Therefore, the data fusion system has to scale for hundreds of vehicles. In addition, the number of targets is not known in advance and our fusion must be robust with respect to clutter and detection failures. Conventional filtering methods that handle each observed vehicle separately, such as multiple Kalman filters or multiple hypotheses tracking~\citep{blackman2004multiple}, require to explicitly solve a complex association problem between the system's sensor detections and tracked vehicles. This severely limits scalability. For this reason, we use the random finite set (RFS) framework~\citep{mahler2007statistical, mahler2014advances}, specifically the Gaussian mixture probability hypothesis density (GM-PHD) filter~\citep{vo2006gaussian}. This filter avoids the explicit data association step and has proven to balance our runtime and scalability constraints well. Additionally, it handles time-varying target numbers, clutter and detection uncertainty within the filtering recursion. 

We add tracking capabilities to our GM-PHD filter by extending it with ideas taken from \citet{panta2009data}. In particular, we make use of the tree structure that naturally arises in the GM-PHD filter recursion and appropriate track management methods. With the resulting tracker we track the measurements of each sensor in parallel. For sensor and motion models, we use a zero-mean Gaussian white noise observation model and a standard constant velocity kinematic model, respectively. The constant velocity model is still a competitive prediction model, while being fast to evaluate~\citep{schoeller2020constant}. All parameters for our sensor and scenario specifications were tuned empirically.

To fuse the tracked data from different sensors and measurement points, we adapt the method from \citet{vasic2015collaborative} that is based on generalized covariance intersection~\citep{mahler2000optimal}. In order to ensure scalability and easy extension of our system setup, we implement a hierarchical data fusion concept, in which we first perform independent local sensor fusion at each measurement point leading to vehicle tracklets. Second-level fusion of all measurement points is then performed on the backend. This step generates the consistent model of the whole highway scene covered by our system, that we refer to as the digital twin.

Switching between different fusion set-ups is possible, depending on the sensors that should be used. Apart from fusing all sensors, there are the possibilities to only fuse the cameras or to only fuse the radars. With this, the system can be adapted to different situations like changing lighting conditions, where the different sensor types complement each other in different proportions. At night for example, our system switches to only using the radars.

\subsection{Position Refinement}
\label{sec:position-regression}

After the data fusion, the detection positions of the vehicles in the digital twin tend to be placed either towards the front or the rear of the vehicles. This is due to the camera detections being two-dimensional bounding boxes and our method for obtaining vehicle positions in world coordinates by casting rays through them and intersecting them with the street plane (see~\sect{sec:object-detection}). The exact placement is dependent on the cameras' perspectives. Other systematic positioning errors our system is affected by are calibration inaccuracies and approximation errors of the street geometry.

To address these problems jointly, we apply a regression-based position refinement using a feed forward neural network after our data fusion. It receives the position of a vehicle in the digital twin and predicts the systematic offset to its actual center position, based on its current location on the highway. By adding this offset to the input position, we get a better estimate of the vehicle's true position and reduce the systematic positioning errors that we described. We have trained the neural network using a randomly split part of our ground truth data, which we associate with vehicle detections from our fusion to obtain a training dataset (see~\sect{sec:groundtruh-generation} and~\sect{sec:evaluation-methodology} for more details). We train one network for operation during the day and one for the night. The networks share the same architecture and have five hidden layers with 32 neurons each. We train them with the Adam optimizer~\citep{kingma2014adam}, batch size 16, learning rate 0.001, and 0.1 weighted L2 regularization for 1000 epochs.

\section{Digital Twin and Extension of Vehicular Perception}
\label{sec:digital-twin}
\begin{figure}[t]
\centering
\subfloat[Camera Image]{\includegraphics[trim=0 3 0 0,clip,width=0.49\columnwidth]{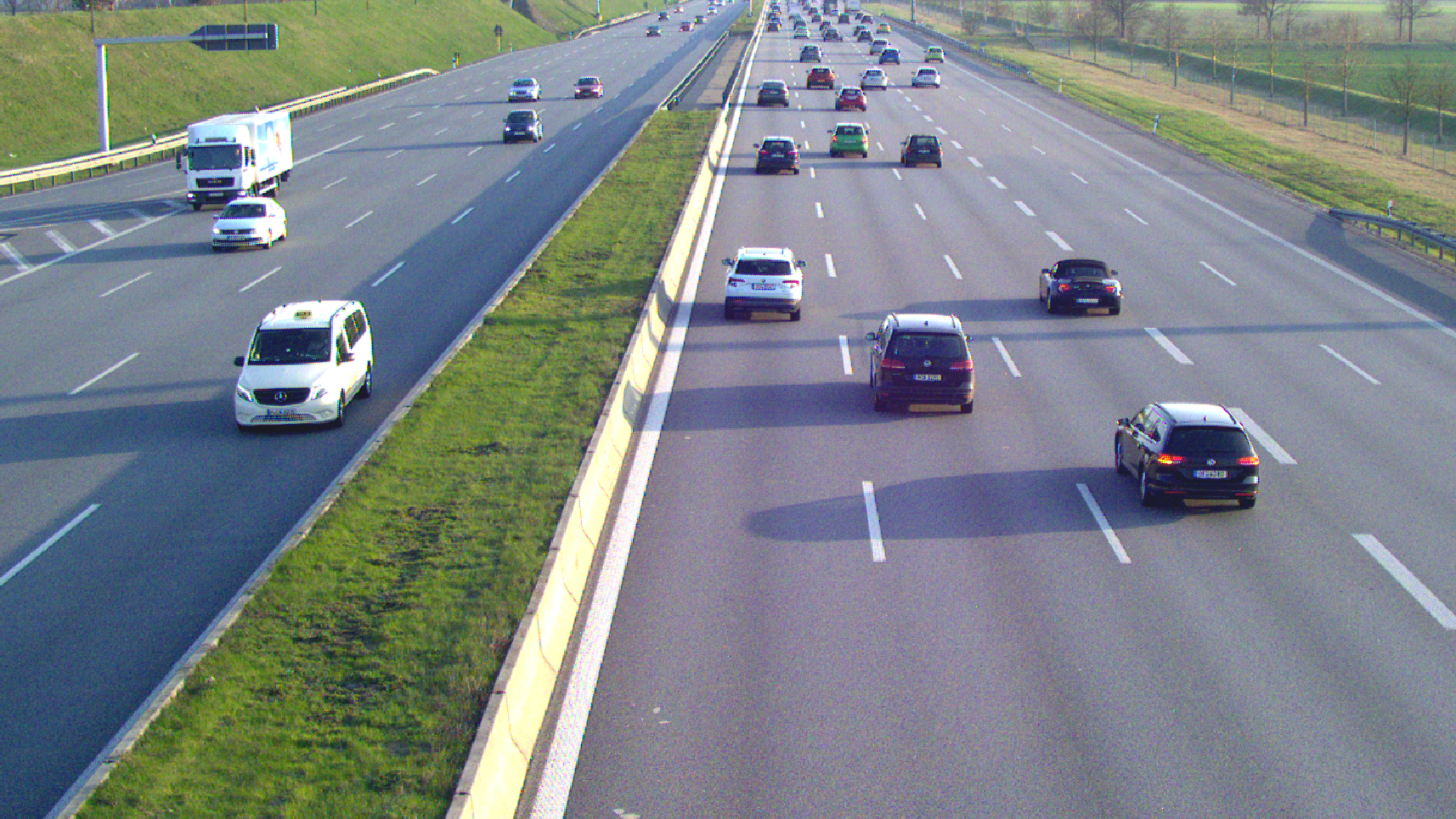}}
\hfill
\subfloat[Digital Twin]{\includegraphics[trim=0 3 0 0,clip,width=0.49\columnwidth]{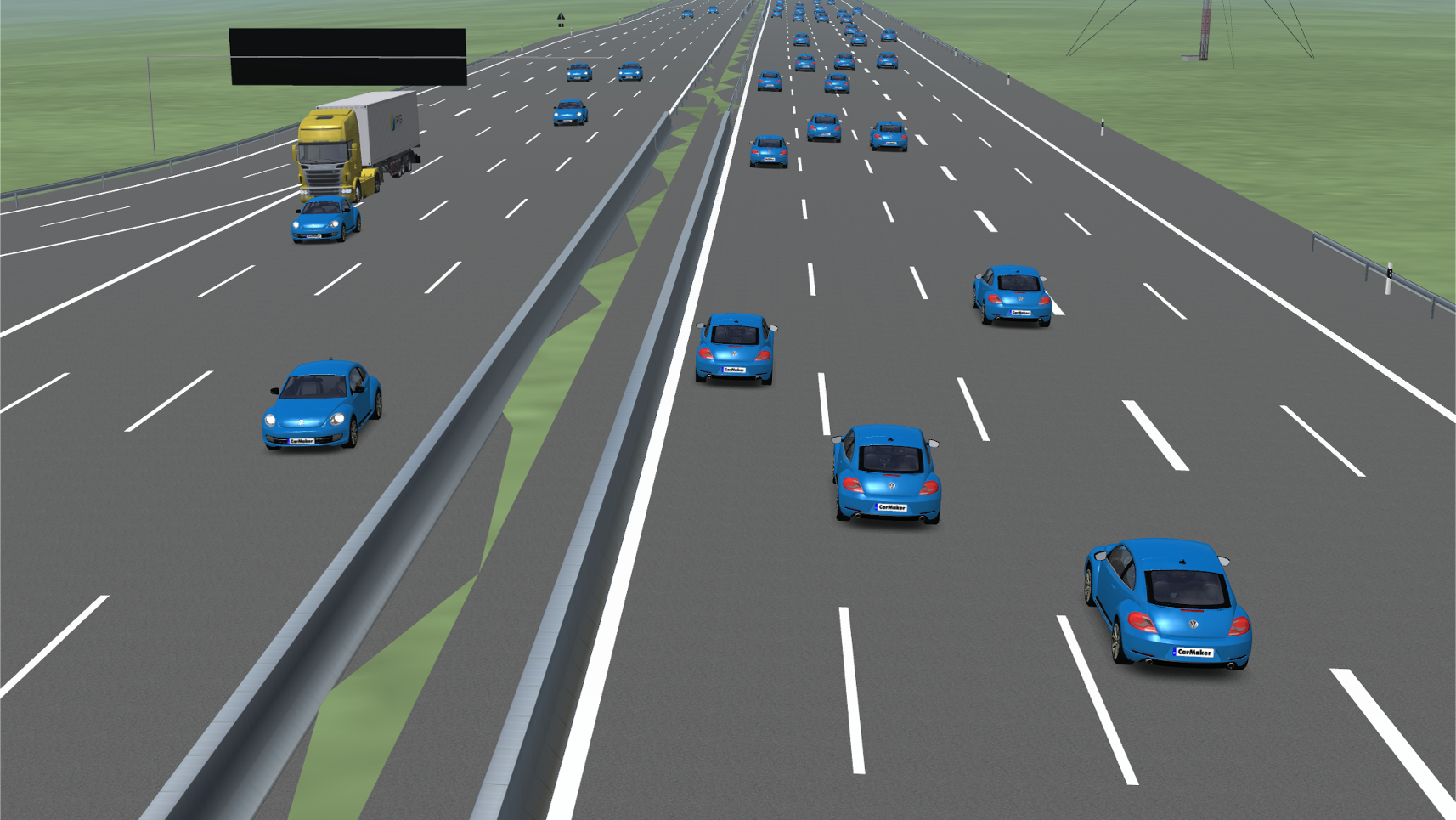}}
\caption{Qualitative example of how our system captures the real world~(a) in a digital twin~(b). We recreate the scene with generalized models for different vehicle types for visualization purposes. During operation, all information is sent to the autonomous vehicle in form of a sparse object list.}
\label{fig:digital-twin}
\end{figure}%

The digital twin represents the main output of the Providentia system. It consists of the position, velocity, and type of every vehicle observed, with each one assigned a unique tracking identifier. The contained positions can be transformed to GPS coordinates, depending on the application the digital twin is communicated to. It can be used by vehicles on the highway stretch to improve their decision making and to implement additional services that can be provided by the infrastructure itself. Such services might include motion prediction for each vehicle, congestion recognition, lane recommendations, and collision warnings. In this section, we illustrate our system's ability to capture the traffic and demonstrate its potential for extending an autonomous vehicle's perception of the scene.

Our qualitative examples were captured in our testbed under real-world conditions. Currently, our system redundantly covers a stretch of about \unit{440}{m} of the road, which corresponds to the distance between the two measurement points. \fig{fig:digital-twin} shows an example of a digital twin of current traffic on the highway as computed by our system. It is a visualization of the information that is also sent to autonomous vehicles to extend their perception. Our system is able to reliably detect the vehicles passing through the testbed. This is only possible by fusing multiple sensor perspectives. The update rate for the digital twin depends on the fusion-setup and the type of the used object detection network. It varies between \unit{13.1}{Hz} when only using the radars at night and \unit{24.6}{Hz} when using the cameras with the YOLOv4 architecture. 

\begin{figure*}[t]
\centering
\includegraphics[trim=150 38 250 400,clip,width=0.98\textwidth]{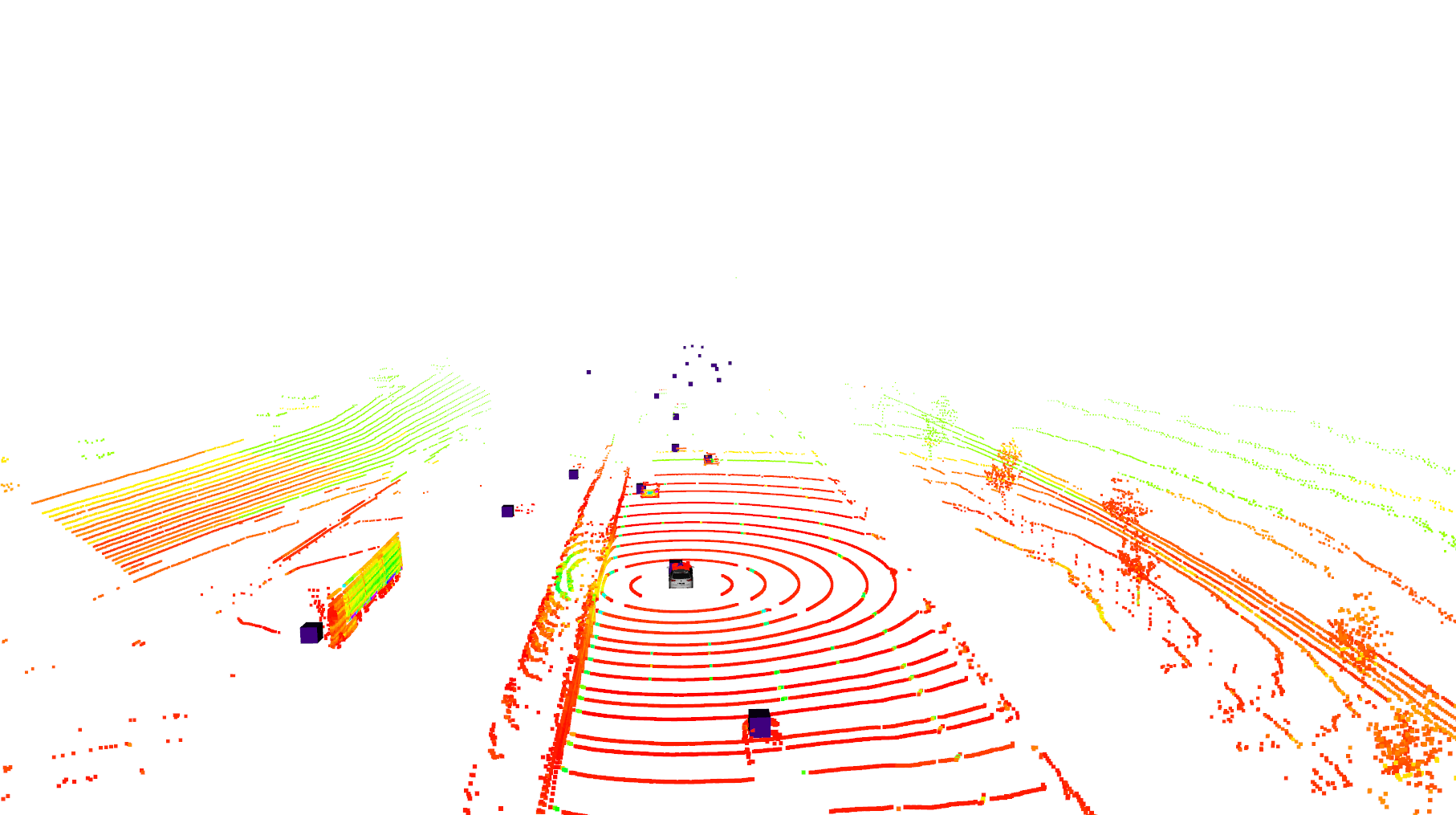}
\caption{An autonomous driving research vehicle driving through our testbed. The dots visualize the vehicle's lidar measurements and the purple cubes represent the vehicles perceived by the Providentia system. While the vehicle's own lidar range is severely limited, its perception and resulting scene understanding are extended into the far distance using information from our system.}
\label{fig:extended-perception}
\end{figure*}%

We transmitted this digital twin to our autonomous driving research vehicle~\textit{fortuna}~\citep{kessler2019bridging} for the purpose of extending its environmental perception and situation understanding. Vehicles perceive their environment by means of lidars, which have limited measurement ranges and the point cloud density in the distance becomes increasingly sparse. Vehicular cameras can capture a more distant environment than lidars are able to, but objects that are too far away appear small on the image and cannot be reliably detected. Furthermore, the vehicle's low perspective is prone to severe occlusions. \fig{fig:extended-perception} illustrates how an autonomous vehicle driving through our system perceives its environment. The violet cubes represent vehicles detected by our system. We observed that the point cloud density of our vehicle's lidars drops significantly at a distance of approximately \unit{80}{m}, but our system's digital twin extends the vehicle's environmental perception to up to \unit{400}{m}. In principle, a system such as ours is able to extend the perception of a vehicle even further, since we designed it with scalability in mind. The maximum distance is only limited by the existing number of built-up measurement points.

\section{Evaluation of the Digital Twin}
\label{sec:evaluation}
To decide what applications can be realized with our system, it is crucial to know the accuracy and detection rate of the digital twin that it generates. For example, using the digital twin for maneuver planning in autonomous vehicles requires high position accuracy, whereas position accuracy is less important for the detection of traffic jams. Knowing the statistical certainty and uncertainty of the system's measurements also makes it possible to define safety margins that vehicles have to take into account when using the provided information.

However, the evaluation of the system's digital twin is a challenging task~\citep{kraemmer2020vorausschauend}. To merely evaluate the detection performance of individual sensors is insufficient to judge the system's performance, as the calibration between the sensors and the fusion algorithms are of paramount importance for the quality of the digital twin and must therefore also be included in the evaluation. End-to-end evaluation of the system requires ground truth information of the traffic on the testbed over an extended period of time. This implies having the exact positions of all the vehicles on the observed stretch of the highway. Labeling the images from the cameras within the system is not sufficient, as it would only provide ground truth information in image coordinates but not in the real world. Using a single, localized test vehicle also has limits, as the system must be able to handle a wide variety of vehicle colors and shapes. Furthermore, the usefulness of simulations is limited as well. In reality, the system is subject not only to various lighting and vibration effects, but also to the decisions of drivers, which are hard to model.

That is why we approximate the required ground truth by recording aerial images of the testbed. These have an ideal -- almost orthogonal -- top-down perspective of the highway. This perspective avoids all inter-vehicle occlusions, and due to their regular contours, vehicles are easy to detect and distinguish. In this section, we will describe how we captured and processed these images to generate ground truth data suitable for evaluating our system. We also explain the evaluation itself in detail and discuss the results together with their implications for the performance of our system.

\subsection{Ground Truth Generation}
\label{sec:groundtruh-generation}

\begin{figure}[t!]
\centering
\includegraphics[trim=0 15 0 0,clip,width=0.9\textwidth]{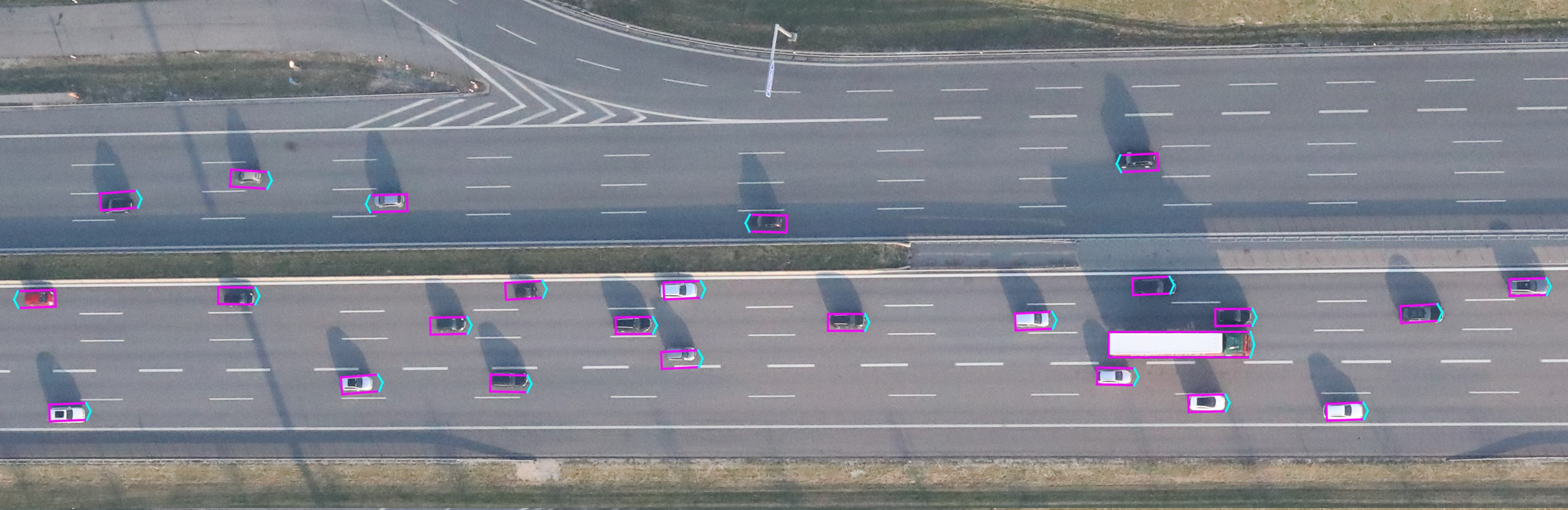}
\caption{Crop of an aerial image including vehicle detections, taken with the helicopter's left-hand camera that captures part of our testbed.}
\label{fig:aerial-highway-image}
\end{figure}%

As previously outlined in~\citet{kraemmer2020vorausschauend}, to generate such an aerial view ground truth, the Providentia testbed was recorded using a 4k camera system mounted on a H135 helicopter. Both the camera system and the Providentia system were synchronized with GPS time. The 4k camera system consists of three Canon EOS 1D-X Mark II cameras oriented in different viewing directions, each recording images at a resolution of 20.2 megapixels. The cameras to the left and right covered the northern and southern parts of the testbed respectively with an overlapping FoV. The third, nadir-looking camera of the system was not used. The cameras captured images simultaneously at a rate of one image per second at a flight altitude of \unit{450}{m} above ground, covering an area of \unit{600}{m}$\,\times \,$\unit{250}{m}. With a focal length of \unit{50}{mm}, each image pixel corresponds to \unit{6}{cm} on the ground. In addition, an IGI Compact MEMS GNSS/IMU system was used to estimate the position and orientation of the sensors during flight to enable georeferencing of the images captured. To optimize the georeferencing accuracy, bundle adjustment with tie points and ground control points was performed.

We use a neural network that has been trained and evaluated with the EAGLE dataset~\citep{dlr-mvda} for the object detection in all aerial images. To compute the positions of the detected vehicles on the road, we cast rays through all four bounding box corner positions in the aerial image and intersect them with a lidar terrain model of the highway surface to compute local UTM coordinates. To obtain the final ground truth data that we use in our evaluation, we compute the center position of each vehicle. We filter out all vehicles that are detected outside of the Providentia system's field of view or are detected twice in the overlap of the two camera FoVs.

The quality of the obtained ground truth depends on the accuracy of the object detections in the aerial images and the positioning accuracy. The detection quality is the subject of on-going research~\citep{azimi2019skyscapes} and not entirely perfect yet. The network we used occasionally missed vehicles, especially trucks, slightly misplaced bounding boxes or detected false positives. We manually corrected all of these errors by re-labeling the concerned bounding boxes to obtain a perfect ground truth. The positioning accuracy in world coordinates depends on the georeferencing accuracy of the images. Specifically, it depends on the calibration accuracy of the 4k system, the quality of the tie and ground control points used for bundle adjustment, and the accuracy of the underlying terrain model. The overall absolute accuracy on the present dataset lies in the centimeter range. The accuracy of this georeferencing is demonstrated in~\citet{kurz2019preciseaerialimage}.

The recording of the ground truth data was performed during the day with a medium traffic volume. \fig{fig:aerial-highway-image} shows a captured aerial image with vehicle detections. Even though the testbed was not always fully covered by the helicopters' cameras, we captured enough vehicles to perform a reliable and statistically significant evaluation. In total, we generated over $2$ minutes of ground truth data containing 2125 valid vehicle observations within our testbed. Additionally, each detection contains a classification that distinguishes between cars and trucks. In total, our dataset contains \unit{95}{\%} cars and \unit{5}{\%} trucks.

\subsection{Evaluation Methodology}
\label{sec:evaluation-methodology}

To compare our digital twin with the ground truth data, we first transform the aerial detections from UTM into the Cartesian world coordinate system used by Providentia, in which the digital twin is defined. In the next step, we match all ground truth frames, i.e. all detections in each pair of left-hand and right-hand camera images, with those frames of the digital twin with the closest corresponding timestamp. Note that the digital twin has a higher frequency than the ground truth with \unit{1}{Hz}, and they are slightly offset with respect to each other. We account for the time difference between matched frames by extrapolating the Providentia detections with a constant velocity prediction. 

\begin{figure}[t]
\centering
\includegraphics[width=1.0\textwidth]{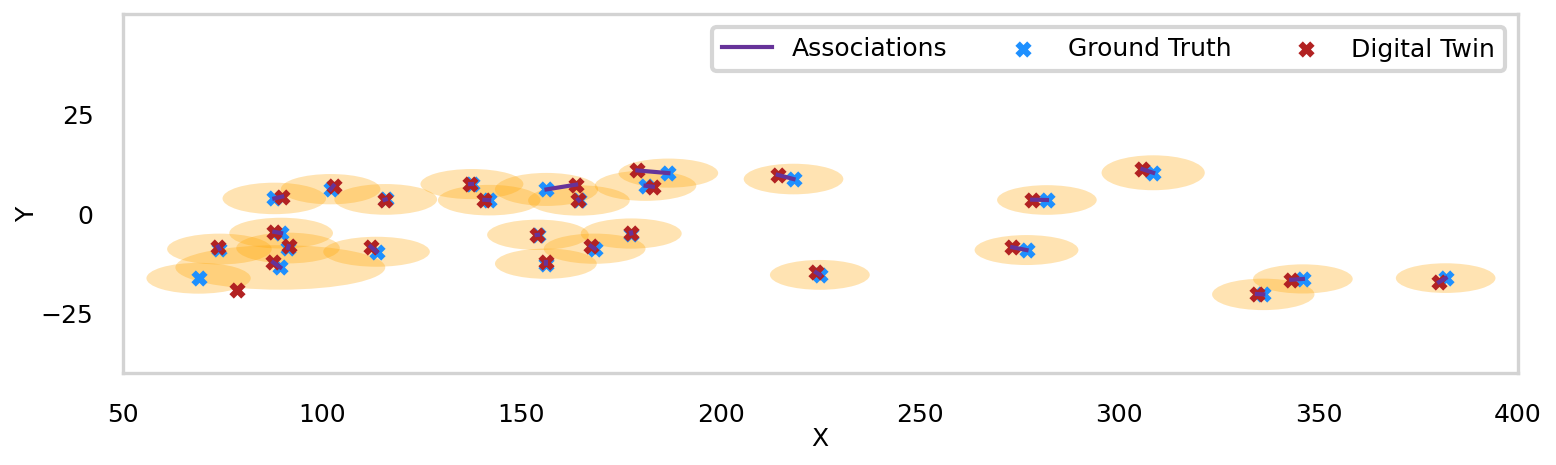}
\caption{Associations between detections from our system and ground truth data. Only detections within the displayed ellipses around each vehicle in the ground truth data are associated. Actual associations are marked with a line between corresponding detections. Note the different size of the ellipses depending on the ground truth vehicle size.}
\label{fig:associations}
\end{figure}%

To assess the overall performance of our system, we evaluate the classification accuracy, the spatial accuracy and the detection rate of the digital twin. For computing these metrics, it is necessary to establish associations between the detected vehicles in the digital twin and the ground truth. For this purpose we use the Hungarian algorithm~\citep{munkres1957algorithms}, which guarantees an optimal one-to-one assignment. We use a weighted Euclidean distance for computing the cost matrix, putting more emphasis on the longitudinal distance between vehicles than on their lateral distance. This accounts for the fact that the estimates in the digital twin have a greater variance in the driving direction because of their shape and kinematics (see also the RMSE results in~\sect{sec:results}). While vehicles driving on nearby lanes can be close, but the estimate of one should not be associated to the ground truth of the other. Furthermore, we make the weights dependent on the vehicles' dimensions, because the detection of a larger vehicle is more likely to be placed further away from its center. In particular, we define the longitudinal weight as the vehicle length plus \unit{8}{m} and the lateral weight as its width plus \unit{1.7}{m}. We determined these additive components empirically. Thresholding this weighted distance represents an ellipse around the ground truth object, with the major axis aligned with the driving direction (see~\fig{fig:associations}). Based on these ellipses, we also perform a gating with threshold 1, in which all associations outside of the respective ellipse are rejected to avoid wrong associations between pairs of false negatives and false positives. Overall, these parameters lead to accurate and reasonable associations.

With the correctly established associations, we can compute our system's classification accuracy. This is the percentage of vehicles that our system assigned the correct class label to. We differentiate between cars and trucks as these are the class labels of the objects contained in our ground truth data. Besides an overall classification accuracy, we also report our system's classification accuracy for both classes individually. Furthermore, we also use such a class-dependent evaluation in the following metrics.

To evaluate the spatial accuracy of the digital twin, we compute the root-mean-square error (RMSE). It represents the standard deviation of the error between the vehicle positions in the digital twin and the ground truth positions in meters, and is therefore a good summary measure of the positioning errors in the digital twin. In particular, for computing the RMSE we split the established associations into a \unit{75}{\%} testset and \unit{25}{\%} training dataset. We use the split training dataset for training our position refinement network (see~\sect{sec:position-regression}), and use \unit{2}{\%} of the training dataset for validation. We make sure that the smaller subset of trucks is split with equal percentages. This helps to avoid skewed outcomes of the random splitting, like the testset containing no trucks.

In addition to its classification and spatial accuracy, the detection rate of our system is important for evaluating its overall performance. Appropriate metrics for this are precision and recall. The precision of our system states which percentage of vehicles detected by our system were actually present. Its recall refers to the percentage of vehicles in the testbed that were successfully detected by our system. To evaluate these detection metrics consistently with each other and the RMSE computation, we compute the true positives, false positives and false negatives based on all established associations. In particular, we first associate all ground truth detections with the digital twin. Then we determine the FoV of our Providentia system and count all associated ground truth vehicles within this FoV as true positives. Those ground truth detections within this FoV that got not associated are false negatives. To compute the false positives, we have to consider that the helicopter was moving and occasionally only covered parts of our testbed to not count correct vehicle detections in our digital twin as false positives because they could not be captured by the ground truth. Hence, we project the current camera FoV of the ground truth data for each frame on the Providentia testbed and intersect it with the Providentia FoV. Then, we take all Providentia detections within this resulting intersected FoV that have not been associated to a ground truth detection as false positives.

We evaluate the performance of the Providentia system at day as well as at night by evaluating the digital twin it creates with either only using camera detections as inputs, or with only using radar detections, respectively. Using only radars is a valid method for simulating night measurements, since radar performance is independent of lighting conditions. Hence, the radar-only based digital twin performs the same way at day and at night, given the same traffic. Like this we can use the same traffic scenes that we recorded during the day for our day and for our night evaluation. In both types of evaluation, we consider the area enclosed by the two measurement points that we cover redundantly.

\subsection{Results}
\label{sec:results}

\begin{table}[t]
\renewcommand{\arraystretch}{1.3}
\caption{Results of the Evaluation of the Providentia Digital Twin}
\label{table:results-all}
\centering
\vspace{0.1cm}
\begin{tabular}{c|c||c||c|c|c||c|c}
\textbf{Mode} & \textbf{Class} & \textbf{Classification} & \textbf{RMSE} & \textbf{RMSE}$_x$ & \textbf{RMSE}$_y$ & \textbf{Precision} & \textbf{Recall} \\ \hhline{========}
\multirow{3}{2em}{\textbf{Day}}     & \textbf{Total} & \unit{96.2}{\%} & \unit{1.88}{m} & \unit{1.81}{m}  & \unit{0.49}{m} & \unit{99.5}{\%}  & \unit{98.4}{\%}  \\ \cline{2-8}
                                    & Car & \unit{96.0}{\%} & \unit{1.79}{m} & \unit{1.72}{m} & \unit{0.48}{m} & \unit{99.6}{\%} & \unit{98.5}{\%} \\ \cline{2-8}
                                    & Truck & \unit{100.0}{\%} & \unit{3.12}{m} & \unit{3.05}{m} & \unit{0.66}{m} & \unit{97.2}{\%} & \unit{96.4}{\%} \\
\hhline{========}
\multirow{3}{2em}{\textbf{Night}} & \textbf{Total} & \unit{95.6}{\%} & \unit{2.00}{m} & \unit{1.82}{m}  & \unit{0.83}{m} & \unit{99.0}{\%}  & \unit{94.1}{\%} \\ \cline{2-8}
                                & Car & \unit{98.2}{\%} & \unit{1.54}{m} & \unit{1.31}{m} & \unit{0.81}{m} & \unit{99.0}{\%} & \unit{93.9}{\%} \\ \cline{2-8}
                                 & Truck & \unit{50.5}{\%} & \unit{5.64}{m} & \unit{5.55}{m} & \unit{1.01}{m} & \unit{98.0}{\%} & \unit{97.1}{\%} \\
\hhline{========}
\end{tabular}
\end{table}

All results of the our evaluation are summarized in~\tab{table:results-all}, separated by day and night as well as by vehicle classes. In the following, we discuss the performance of our system for all evaluated metrics in detail. 

\textbf{Classification Accuracy.} 
Our system classifies \unit{96.2}{\%} of the vehicles correctly during the day and \unit{95.6}{\%} during the night. At day, trucks are perfectly classified, while the accuracy for cars is \unit{96.0}{\%}. Most classification errors stem from vans, that are assigned to the car class in our ground truth, but get easily confused with trucks by our camera detection network. When such misclassifications in the camera detections happen consistently to a vehicle, our tracker cannot compensate the error. However, these errors could be reduced by retraining our detection network with an additional van class, such that it learns to better differentiate. At night, using our radars, our system has a high classification accuracy of \unit{98.2}{\%} for cars, but struggles at classifying trucks. They are mistaken for cars half of the time. However, this is expected, because the classification of the radars is only based on reflection characteristics and coarse length estimates, but no visual clues can be used.

\begin{figure}[t]
\centering
\captionsetup[subfigure]{margin={0.5cm, 0cm}}
\subfloat[Errors at Day]{\includegraphics[trim=0 3 0 0,clip,width=0.49\columnwidth,valign=t]{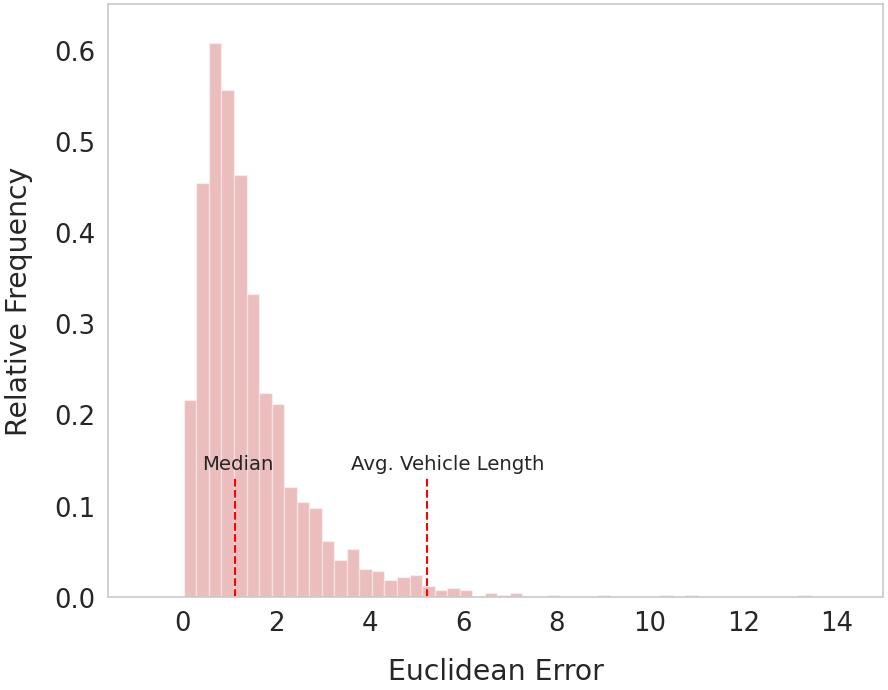}}
\hfill
\subfloat[Errors at Night]{\includegraphics[trim=0 3 0 0,clip,width=0.49\columnwidth,valign=t]{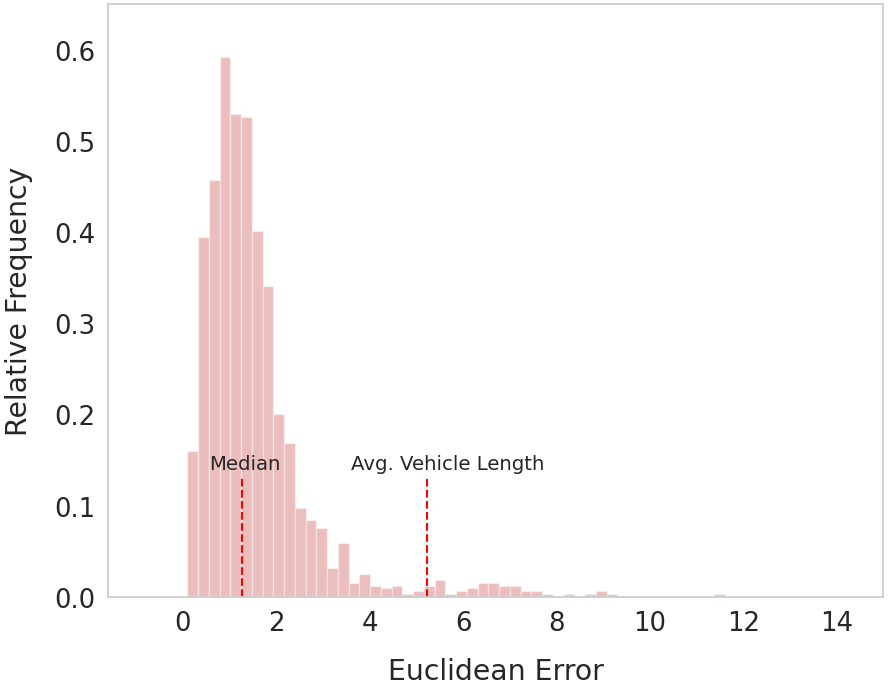}}
\caption{Distribution of positioning errors in the digital twin compared to the aerial ground truth. Our system is very accurate in most cases. The errors are within the average vehicle length in \unit{98.0}{\%} of the cases at day and in \unit{99.6}{\%} of the cases at night. In \unit{50}{\%} of the cases the error is less than \unit{1.02}{m} at day and less than \unit{1.26}{m} at night.}
\label{fig:error-distribution}
\end{figure}

\textbf{Spatial Accuracy.} Concerning the spatial accuracy of the digital twin, we achieve an overall RMSE of \unit{1.88}{m} during the day and \unit{2.00}{m} at night (see~\tab{table:results-all}). In both cases, the major component of the RMSE stems from the longitudinal direction and is \unit{1.82}{m} and \unit{1.81}{m}, respectively. In the lateral direction, our system is very precise with an error of~\unit{0.49}{m} during the day. At night, this error component increases to~\unit{0.83}{m} because the radars' lateral position estimates are subject to greater noise, especially at larger distances. In both cases, the high lateral accuracy allows us to reliably determine the lane for each vehicle.

A large component of both the longitudinal and lateral positioning errors is due to the current lack of information about the actual extents of the objects in our system. Because our camera detections are two-dimensional bounding boxes in the image plane, the vehicles' lengths are not explicitly taken into account for estimating their position, and their widths can only be approximated with perspective errors. The radars on the other hand provide estimates of the vehicles' centers, that result from the detection point which is corrected by average vehicle extents. These average extents are associated with the corresponding vehicle class that is inferred from its reflection characteristics. We partially compensate this source of systematic errors for both sensor types with our position refinement regression, but cannot eliminate it completely without knowing the vehicles' exact extents. Therefore, our system has difficulties handling vehicles which extents deviate from the average of their respective vehicle classes. As a result, our system has bias that tends to place detections more towards the rears or towards the fronts of the vehicles, depending on the perspective. However, in the ground truth the position of an object is specified at its center. As the ground truth vehicles driving through the testbed over the course of our evaluation have an average length of approximately \unit{5.18}{m}, the extreme placement of a detection at the rear or front would already cause a displacement of \unit{2.6}{m} to the center. Taking this into account, our spatial accuracy is very promising at both day and night.

\fig{fig:error-distribution} shows the distributions of our overall positioning errors from which we computed the RMSE. During the day, in \unit{50}{\%} of the cases our error is less than \unit{1.10}{m} and at night less than \unit{1.23}{m}. Furthermore, in \unit{98.0}{\%} respectively \unit{99.6}{\%} of cases the errors of our detections are within the average vehicle length. For trucks the error is in all cases within the average truck length. This indicates that by incorporating the vehicles' extents from the sensors, e.g. by computing three-dimensional bounding boxes, our positioning errors could be further reduced.

\begin{figure}[t]
\centering
\subfloat[Errors at Day]{\includegraphics[trim=0 3 0 0,clip,width=0.98\columnwidth]{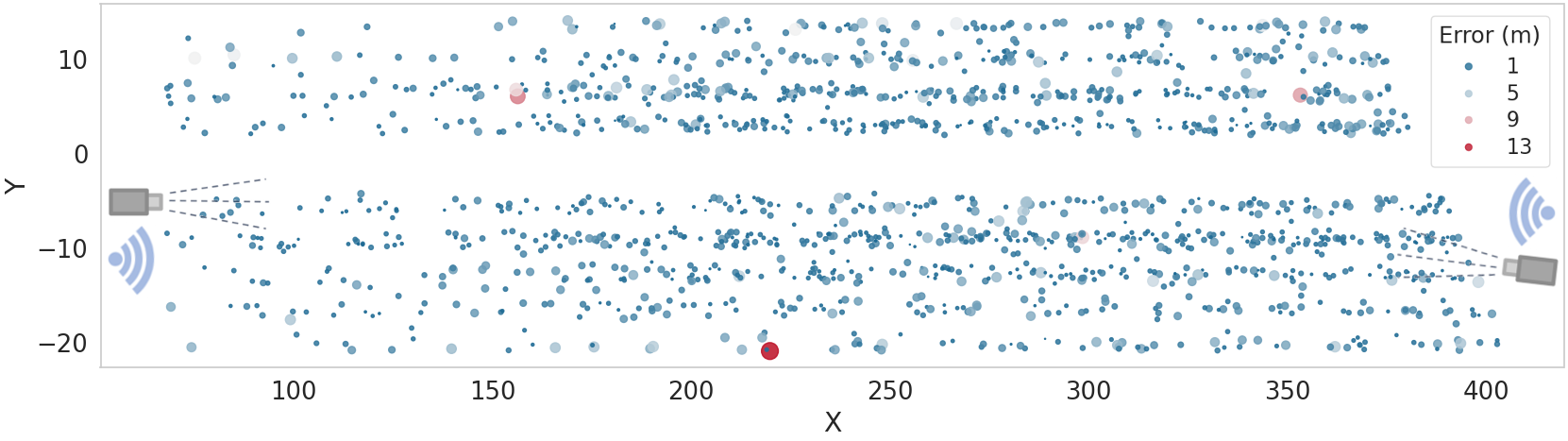}}
\vfil
\subfloat[Errors at Night]{\includegraphics[trim=0 3 0 0,clip,width=0.98\columnwidth]{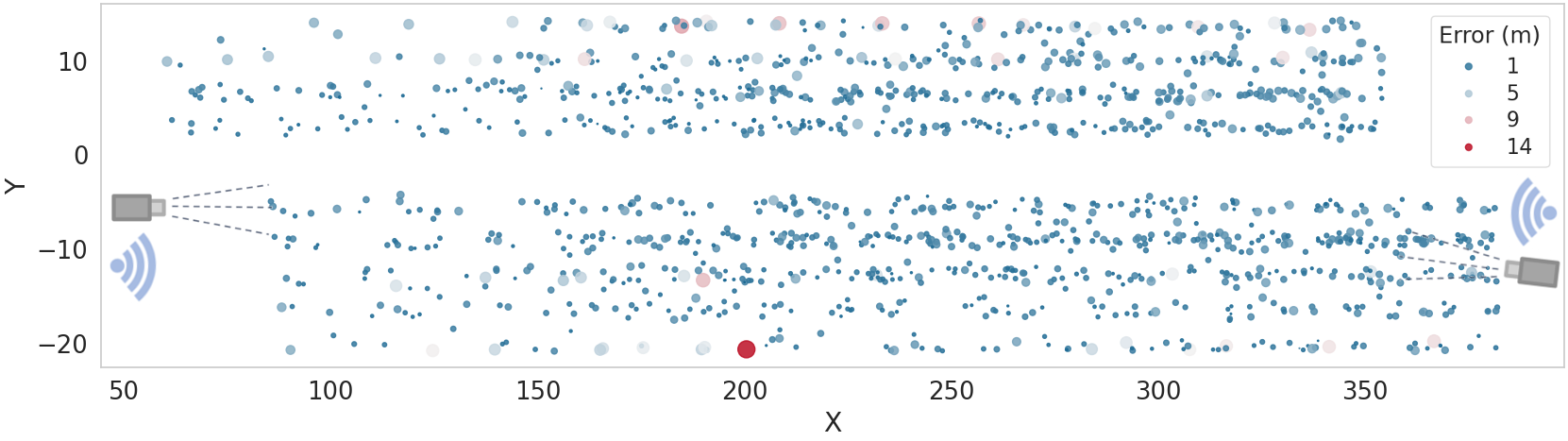}}
\caption{Positioning errors in the digital twin for each ground truth vehicle on the highway, along with the schematic measurement point positions. The FoV differences between day and night are a result of the changing sensor setups. Both during the day and at night severe errors are rare, and mostly due to oblique perspectives and greater vehicle extents.}
\label{fig:error-map}
\end{figure}

In general, by separating the the positioning errors for both vehicle classes, in \tab{table:results-all} it can be seen that the RMSE for cars is smaller than for trucks at day as well as at night. Based on our previous analysis, this is expected, since trucks have greater extents that lead to larger estimation errors for the vehicle centers. For trucks, our system performs significantly better at day than at night. This can be explained by the radar's high misclassification rate for trucks. A consistently false classification of a truck as a car leads to an underestimation of the vehicle's center offset from the detection point. For cars, our system performs slightly better at night because of smaller longitudinal errors in the radar detections and the radars' center correction performing well for cars.

To see how the positioning errors are distributed over our testbed, in \fig{fig:error-map} we plotted the errors to each corresponding ground truth detection. At day and at night, large positioning errors over \unit{5}{m} are rare and often belong to large vehicles. Additionally, larger positioning errors mostly occur on the top and bottom lanes. There, the sensors have more oblique perspectives on the vehicles and the raycasting through the lower-edge midpoints of the bounding boxes deviates more from the actual middle of the vehicles in the camera detections. The radars are more accurate at measuring the positions of vehicles that drive closely to their viewing direction, as their angular resolution deteriorates towards lager angles. Furthermore, the road is only approximately a plane and deviates the most from it towards the outer lanes. Hence, there the projection errors for both the camera detections and the radar detections are the greatest.

A large number of systematic positioning errors are corrected by our position refinement module. Even though small errors caused by oblique camera perspectives remain, they were significantly reduced. Without the regression we also had an accumulation of errors for the camera positioning towards the middle of the testbed. In this area, the vehicles are far from all cameras and their resolution in the images is the smallest, which results in higher uncertainty in the bounding box estimates. The way we compute the vehicle positions from the camera detections, i.e. by intersecting rays with the road (see~\sect{sec:object-detection}), is sensitive to such inaccuracies over large distances, because the rays intersect the ground plane at a flatter angle. Our radars had the largest errors for the vehicles on lanes of the opposite driving direction of where they were installed, which the regression also addressed well. In total, the position refinement module reduced the RMSE during the day from \unit{3.47}{m} to \unit{1.88}{m} and at night from \unit{2.64}{m} to \unit{2.00}{m}. While the data-driven correction has a significant effect on errors at both day and night, the camera detections benefit stronger from it. This shows that the cameras suffer more from systematic errors, for example caused by unfavourable perspectives in some areas of the road. Furthermore, the smaller error reduction for the radars can be explained by their correction of the center offset with average vehicle extents, and it shows the need for this correction in the camera detections. Despite the significant improvements, some random error sources could not be fully corrected in both the data fusion and regression, for example measurement noise and sensor vibrations that make the detections susceptible to calibration inaccuracies.

\textbf{Detection Rate.} Lastly, evaluating the detection rate, our system achieves an overall precision of \unit{99.5}{\%} during the day and \unit{99.0}{\%} at night (see~\tab{table:results-all}). This means that we have very few false positives and almost all vehicles detected by our system do actually exist and are correct. The few false positives during night can be explained with the radars tending to split larger trucks in two detections that are classified as truck or car. In most cases, this splitting is compensated by our tracker, but it is not always possible. Quite similarly, larger trucks are sometimes split into towing vehicle and trailer by the camera detections explaining the slightly lower precision for the truck class. As for the recall, our system achieves \unit{98.4}{\%} during the day. Hence, we detect \unit{98.4}{\%} of all ground truth vehicles on the highway and only miss \unit{1.6}{\%} of them. At night, our recall is \unit{94.1}{\%} which is not as high as during the day, but the vast majority of the vehicles is still detected. The reason for this decrease can be seen when differentiating between cars and trucks. The recall for trucks with \unit{97.1}{\%} compared to \unit{93.9}{\%} for cars is significantly higher. Trucks have a larger surface to reflect radar signals and thus the likelihood of missing a truck is smaller than for a car. During the day, our system is slightly more accurate at detecting cars. Most likely, because cars are more frequent than trucks in the training data of our object detection networks.

It is important to note that we analyzed precision and recall of our system on a frame-by-frame basis. Hence, when we do not detect a vehicle, it does not imply that it is passing through the testbed completely undetected. This did not happen. Rather, at specific moments in time, certain vehicles may be briefly lost due to occlusions caused by larger vehicles. This indicates that incorporating an occlusion handling mechanism in the tracking can further optimize the overall performance.

Overall, our system achieves a high degree of reliability, both in terms of spatial accuracy and detection rate, as well as in classification accuracy. The accurate positioning of our system also allows us to estimate the vehicles' motion directions and speeds with high precision. Our results further show that it is highly beneficial to use cameras during the day instead of a radar-only system. Their detection rate and classification accuracy are higher than that of radars. Furthermore, their update frequency is almost twice as high, along with similar overall spatial accuracy. However, also our radars show good performance that makes our system reliable even at night. Hence, both sensor types complement each other and enable our system to run at any time of the day. By incorporating methods to determine spatial vehicle extents, our system's positioning accuracy could be further improved and a more balanced and larger training dataset could lead to even higher detection and classification performance, thus further increase the system's reliability.

\section{Lessons Learned} 
\label{sec:lessons-learned}
Only technological advances in recent years made it possible to develop a system like Providentia. This especially concerns computing power, artificial intelligence and data fusion algorithms. However, despite these advances it is a complex task to build a functional IIS for fine-grained vehicle perception such as our system.

Prior to building up the actual IIS, we recommend conducting many different field tests to gather a good understanding of the challenges involved. This is necessary due to the diverse nature of problems posed by every region or road. To name a few examples, the selection of appropriate sensors and their mounting locations heavily depend on the length of the road segment, its curvature and the direction of observation. For our research testbed, one important aspect was to ensure that it is as diverse as possible. To achieve this, we chose a highway section at one of Germany's traffic hot-spots that leads into a highway interchange and in addition is equipped with ramps that lead towards a close by city. Therefore, many interesting driving maneuvers take place, for example various lane changes, vehicle interactions, and even accidents sometimes. And our testbed is exposed to various traffic conditions, from light traffic to heavy traffic jams. But also mixtures of light traffic and traffic jams occur, for example at moments when the two lanes that are branching off towards the intersecting highway get congested, while on the inner lanes vehicles are passing with high speeds.

Based on our experiences, building up the hardware for such a system is technically demanding and requires a team with a wide skillset. Not only was it necessary to design tailored brackets to equip the gantry bridges with our sensors, deploy the sensors and computing units in a weather resistant manner, install cabling to get a high-speed internet connection at the highway, but also legal and safety questions had to be answered. However, we perceive the ongoing trend to technologically modernize roads and highways to observe and actively optimize traffic in many countries. In future, this will synergize with systems like Providentia and significantly reduce both costs and effort to build such a system.

After the initial construction, our system has been running for over two years at the time of writing and its hardware required only little maintenance. What is more difficult is the system's calibration, because the high number of multimodal sensors results in many degrees of freedom. Furthermore, even after a precise initial calibration, the system gradually decalibrates itself over time. This is primarily caused by temperature changes and oscillations of the measurement points due to wind and vibrations caused by passing vehicles, especially trucks and buses. All these effects slightly change the sensors' positions with respect to the road over time. Hence, the calibration must be regularly adjusted to avoid performance deterioration. In practice, during our project we re-calibrated our system when inaccuracies became visually apparent. Even though these are only small readjustments once the system has been thoroughly calibrated initially, for future applications we recommend to use suitable auto-calibration methods to reduce maintenance efforts. As the oscillations also cause the gantry bridges to slightly swing around their equilibrium, auto-calibration methods that run online and compensate all immediate oscillations would further improve the system's performance and reliability in the future.

\begin{figure*}[t]
\centering
\includegraphics[trim=15 200 30 0,clip, width=0.8\textwidth]{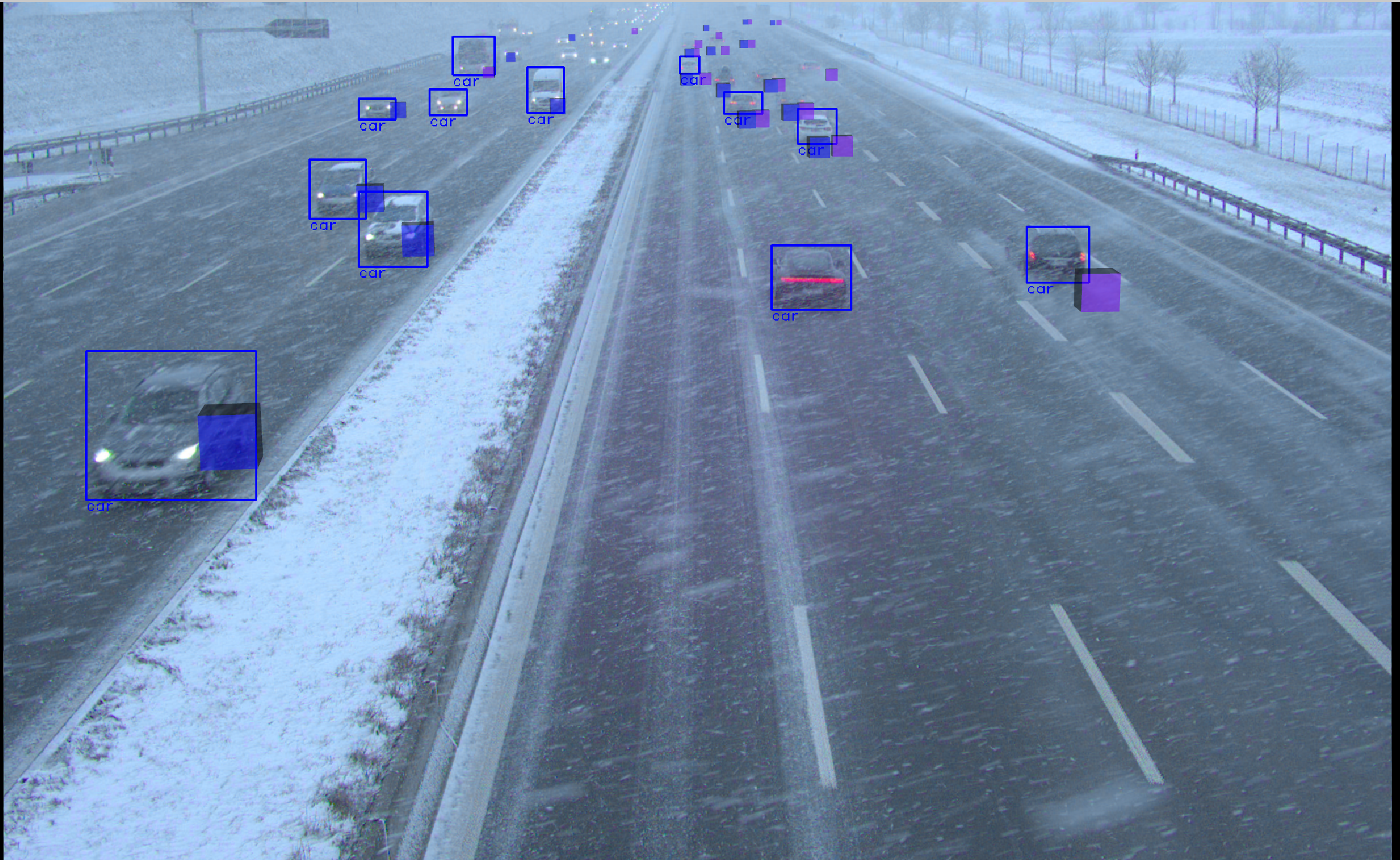}
\caption{Detections of our system in a blizzard. Camera detections are marked with bounding boxes and the detections of our radars with cubes. Our radars detect distant vehicles more reliably under this weather conditions.}
\label{fig:snow-detections}	
\end{figure*}

In this work we extensively evaluated our system with a recorded ground truth. We covered a wide variety of different vehicle types, movement patterns and interactions. This is necessary to answer research questions and to develop our system, such that it achieves high positioning precision and detection rates. However, during long-term operation it would be beneficial to frequently monitor the quality of the system's digital twin, where using a helicopter can get costly. Therefore, methods that trade test coverage and precision for cheaper cost must be developed. Ways to achieve this could be the use of drones or several localized test vehicles, and in future even autonomous vehicles driving through the system's FoV. During our research project, we were able to qualitatively assess the performance of our system in harsh weathers. As shown in~\fig{fig:snow-detections}, even in snow storms our system is able to reliably detect the vehicles passing through our testbed. However, to quantitatively evaluate the system's performance under such conditions and to identify cross-over points at which it is ideal to switch between different fusion modes, in future the development of evaluation methods that do not rely on aerial observation are required additionally. Furthermore, self-diagnosis could instantly detect sensor failures or faults within the fusion system, for example caused by a deteriorated calibration. For this purpose, approaches like the one from \citet{geissler2020plausibilitybased} could be applied and extended. 

Regarding future extension, we have built our system such that it is scalable (see~\sect{sec:architecture}). Many of our developed concepts and algorithms can be transferred when increasing the number of measurement points. However, some further developments are necessary for a distributed system with many measurement points. For example, scaling down the computing units towards being embedded in the sensors could reduce hardware costs. Furthermore, the selection of an appropriate middleware must be given considerable thought and attention. While popular open-source solutions like ROS present a convenient platform and enable a short development time, aspects related to the communication architecture and message transport need to be assessed carefully for delays and bottlenecks that may affect the real-time performance of the scaled system. Typical sources of bottlenecks are the transfer of images through TCP based serialized transport. Another downside is the missing backwards compatibility of ROS messages. Once data has been recorded in an old message format, there are problems replaying it with adapted new messages.

The quality of the entire system depends upon a precise knowledge about the time at which various events occur. We found an appropriate strategy for time synchronization based on a dependable master clock to be essential. While working with off-the-shelf sensors, in future it would be beneficial to choose those which support synchronization with an external master clock.

We see a great potential in not only extending the system on the highway but also in cities. One interesting question to be answered in this context is the density of measurement points needed to support autonomous vehicles. Perhaps, full coverage is not necessary, but the density of measurement points should be increased in dangerous traffic areas, while thinning out in areas with permanent light traffic density.

\section{Conclusion} 
\label{sec:conclusion}
To improve the safety and comfort of autonomous vehicles, one should not rely solely on on-board sensors, but their perception and scene understanding should be extended by adding information available from a modern IIS. With its superior sensor perspectives and spatial distribution, an IIS can provide information far beyond the perception range of an individual vehicle. This can resolve occlusions and lead to better long-term planning of the vehicle.

While there is much research currently being done on specific components and use-cases of IIS, information on building up an entire system is sparse. In this paper we described how a modern IIS can be successfully designed and built. This includes the hardware and sensor setup, detection algorithms, calibration, data fusion and position refinement. We have shown that our system is able to achieve good results at capturing the traffic of the observed highway stretch and that it can generate a reliable digital twin in near real-time. We have further demonstrated that it is possible to integrate the information captured by our system into the environmental model of an autonomous vehicle to extend its limited perception range.

Our extensive quantitative evaluation has shown that our system is characterized by both a high classification and spatial accuracy, as well as a high detection rate, at day and night. The primary purpose of our system is to enhance the perception of autonomous vehicles in the testbed. But based on the results of our evaluation, it is also evident that a system like ours could be used for applications such as traffic prediction, the detection of emerging traffic jams, wrong-way drivers and immobile vehicles. Traffic flow management with lane and speed recommendations could be another possible application. Beyond this, the system could be used as a reference for testing, evaluating and developing autonomous driving functions. And it is a huge data source for developing data-driven algorithms.

We described our experiences with building the Providentia system and outlined some possible improvements, especially regarding its scalability, like adding automatic online calibration and methods for continuous quality monitoring. Furthermore, taking into account the vehicles' spatial extents would improve its positioning accuracy and further reduce errors caused by different camera perspectives. Besides this, in future we plan to make our system more robust in adverse weather conditions as well as during traffic jams with severe occlusions.

\section*{Acknowledgments}
This research was funded by the Federal Ministry of Transport and Digital Infrastructure of Germany in the projects Providentia and Providentia++. We would like to express our gratitude to the entire Providentia team for their contributions that made this paper possible, namely its current and former team members: Vincent Aravantinos, Maida Bakovic, Markus Bonk, Martin B{\"u}chel, M{\"u}ge G{\"u}zet, Gereon Hinz, Simon Klenk, Juri Kuhn, Daniel Malovetz, Philipp Quentin, Maximilian Schnettler, Uzair Sharif, Gesa Wiegand, as well as all our project partners. Furthermore, we would like to thank IPG for providing the visualization software and the Bavarian highway operator (Autobahndirektion S\"udbayern) for their continuous support when building up the infrastructure.

\bibliographystyle{apalike}

\end{document}